\pdfoutput=1

\documentclass[11pt]{article}
\usepackage{hyperref}

\usepackage{acl}
\usepackage{graphicx}

\usepackage{times}
\usepackage{latexsym}

\usepackage[T1]{fontenc}

\usepackage[utf8]{inputenc}

\usepackage{microtype}
\usepackage[symbol]{footmisc}

\usepackage{microtype}

\title{Multilinguals at SemEval-2022 Task 11: Complex NER in Semantically Ambiguous Settings for Low Resource Languages}
\author{Amit Pandey*, Swayatta Daw*, Narendra Babu Unnam\and Vikram Pudi \\
        International Institute of Information Technology, Hyderabad, India\\
        \texttt{\{amit.pandey, swayatta.daw\}@research.iiit.ac.in}\\ \texttt{narendra.babu.unnam@research.iiit.ac.in}\\  \texttt{vikram@iiit.ac.in}}

\begin{document}
\maketitle
\begin{abstract}
We leverage pre-trained language models to solve the task of complex NER for two low-resource languages: Chinese and Spanish. We use the technique of Whole Word Masking (WWM) to boost the performance of masked language modeling objective on large and unsupervised corpora. We experiment with multiple neural network architectures, incorporating CRF, BiLSTMs, and Linear Classifiers on top of a fine-tuned BERT layer. All our models outperform the baseline by a significant margin and our best performing model obtains a competitive position on the evaluation leaderboard for the blind test set.
\footnotetext[1]{Equal Contribution}
\end{abstract}

\section{Introduction}
The Named Entity Recognition (NER) task aims to identify the named entities in an input sequence and categorize them into certain predefined categories or class labels. The task of NER can be further broken down into two subtasks: 1) identification of the entity span and 2) classification of the identified entity span into predefined class labels. 
For example, in the sentence: \textit{New York City is the most densely populated major city in the United States.}, \textit{New York City} is a named entity of type LOCATION with an entity span of 3 tokens.

The most popular NER task in the English language is CoNLL \cite{baevski-etal-2019-cloze}, which is widely used as a benchmark for most NER models. Multiple models have been able to obtain sufficiently high performances in this task setting \cite{ace+document, coregularised_luke, cross-sentence-context, flert, plmarker, luke, clkl}. The CoNLL training set consists of 14,987 train sentences, which comprise 203,621 tokens in total for English data. The entity space consists of 4 different types of entity type labels (locations, persons, organizations, and miscellaneous) to classify each named token. The English data was taken from the Reuters Corpus, which comprises of Reuters News Stories for one year. The training data source, and by extension the labeled named entities, comprises of majorly popular entities found in the general English textual content prevalent in the media. Hence, these entities were easier to classify into the correct classes due to the large prevalence of training data. With the use of pre-trained transformer-based language models, which are already trained on a large unlabelled corpus of English text, this task became even less challenging, as the nature of textual structure in these corpora largely overlap with that of CoNLL. 

However, this task becomes challenging in practical settings where a multitude of varieties of named entities is possible. Many of these entities, like Creative Works (CW) and Products (PROD) have complex and ambiguous textual structural content. Such complex named entities rarely appear even in the large training data sets, and the length and structure of the named entities keep changing. 

We investigate the task of complex, semantically ambiguous, and low-resource NER~\cite{multiconer-report}. This task is based on the complex NER, search query and code-mixing NER challenges introduced by \citet{meng2021gemnet} and \citet{fetahu2021gazetteer}. The shared task of MultiCoNER (stands for multilingual complex NER) adds additional challenge by introducing rarer label types (like Creative Work, Product, etc.). 

Another way to increase the difficulty of NER task is to perform it for low-resource languages. There is a significant dearth of both labeled and unlabelled data for such languages. The complexity is further enhanced by using rarer entity types in such languages. Therefore, the scarcity of training data, along with the rarity of entity types, makes it difficult for the models to perform better in the low-resource setting.
The shared task of MultiCoNER introduces datasets in multiple low-resource languages.

We leverage large pre-trained language models trained on low-resource language corpora to obtain competitive performances in the low-resource and complex NER setting. We show that simpler architectures successfully outperform other heavier counterparts. We use standard BERT+CRF-based models to obtain high performances in the evaluation set. We experiment on two low-resource datasets: Spanish and Chinese.
The code is available at \url{https://github.com/AmitPandey-Research/Complex_NER}

We compare multiple architectures on the test and validation set of the shared task. The organisers provide a baseline of XLM-RoBERTa model \cite{xlm-r} finetuned on the training dataset of the specified language for the given task. In this paper, we treat the finetuned XLM-RoBERTa as baseline (two separate models are trained for Spanish and Chinese language) and compare the performances against our models.  
All our models beat the baseline by a significant margin.
We describe the prior research work done with respect to both general and low-resource NER tasks in Section \ref{Related Work}. We provide the formal task description in Section \ref{taskdescription}, the dataset details in Section \ref{dataset}, the method and the model architecture in Section \ref{method}. We provide details about the experimental implementation in Section \ref{implementationdetails}. We discuss the results obtained and error analysis in Sections \ref{results} and \ref{erroranalysis} respectively, and finally, we conclude the paper in Section \ref{conclusion}.

\section{Related Work}
\label{Related Work}
The task of low-resource NER has been investigated before by multiple researchers. This line of research focuses mainly on leveraging the cross-lingual contextual information obtained from low-resource languages. \cite{10.5555/3304222.3304336} use cross-lingual knowledge transfer to train the NER model for the low-resource target language. \cite{xie-etal-2018-neural} use bilingual dictionaries to tackle the task of low-resource NER. \cite{Rahimi2019MultilingualNT} proposes a Bayesian graphical model approach to improve performance on NER tasks. 

NER models often use gazetteers (list of named entities) to improve performance in NER tasks. \cite{rijhwani-etal-2020-soft} creates soft-gazetteers for low-resource languages, leveraging English Knowledge Bases. \cite{zeroresource} focuses on an unsupervised approach for NER for to circumvent the label scarcity problem in low-resource languages. \cite{rahimi-etal-2019-massively} leverages multilingual transfer learning from multiple languages for low-resource NER tasks. \cite{DBLP:journals/corr/abs-2102-13129} uses distant supervision in the low-resource setting for NER. 

There are multiple approaches that have been undertaken in the recent past to improve the state-of-the-art in NER tasks. \cite{DBLP:journals/corr/abs-2010-05006} uses concatenation of embeddings to outperform the state-of-the-art in NER tasks, as they infer that concatenation of embeddings leads to a better word representation. Their method automates the process of finding meaningful embeddings to concatenate for improved performance. \cite{coregularised_luke} propose a co-regularization framework for entity extraction comprising of multiple models with different architectures but different parameter initializations. This helps to tackle overfitting of large neural network-based models on low-resource training data for NER. \cite{flert} use document-level features to improve information extraction on entity-centric tasks.

NER and Relation Extraction are the core information extraction tasks in NLP. \cite{plmarker} models this as a span-pair classification problem, and they further improve the pair representations by considering the dependencies between the spans (pairs) by strategically packing the markers in the encoder. \cite{luke} proposes a novel entity-aware self-attention framework for transformer based models for NER. \cite{clkl} extracts document-level context for sentences for which document information is absent. They treat the sentence as a query and use a search engine to extract the document level contextual information. \cite{cross-sentence-context} uses multiple neighbouring sentences as the contextual information for NER.

\textbf{Pre-trained Language Models for NER :} Ever since the introduction of BERT~\cite{devlin-etal-2019-bert}, transformer based pre-trained language models have effectively utilized transfer learning for downstream NLP tasks. NER has been traditionally modeled as a sequence labeling problem. \cite{bilstmcrf} proposed a Bidirectional LSTM with a CRF layer on top for classifying tokens as entities. \cite{bertcrf} use a pretrained BERT model with a CRF layer on top for performing NER on the DailyHunt news dataset. We use a BERT-based model with a CRF layer on top and achieve competitive performance on low-resource NER tasks on multiple languages, beating the baseline by a significant margin in each case. 

\section{Task Description}
\label{taskdescription}
In this task, we attempt complex NER for two low-resource languages: Spanish and Chinese. This task presents additional challenges in the form of test instances consisting of short search queries and low-context sentences. For this task, the systems had to identify the B-I-O format \cite{ramshaw1999text} (short for beginning, inside, outside) tags for six NER entity type labels: 1) Person, 2) Product, 3) Location, 4) Group, 5) Corporation, and 6) Creative Work. The description of these labels is shown in Table \ref{tab:stats2}.
\begin{table}[]
    \centering
    \begin{tabular}{c|c}
          \textbf{Label} & \textbf{Description} \\
          \hline
         PER & Person \\
         LOC & Location \\
         GRP & Group \\
         CORP & Corporation \\
         PROD & Product \\
         CW & Creative Work \\
    \end{tabular}
    \caption{Entity types in the label space}
    \label{tab:stats2}
\end{table}

\section{Dataset}
\label{dataset}
The MultiCoNER dataset~\cite{multiconer-data} consists of multiple low-resource languages. We consider Chinese and Spanish languages in this paper. For the monolingual track, the participants have to train a model that works for a single language. We fine-tune the language model on the train set to obtain predictions on dev and test set. The labels from the blind test set are not disclosed. The dataset follows a BIO tagging scheme and there are 6 entity types in the label space. The statistics for the Chinese and Spanish datasets in the monolingual track for the train and dev set are provided in Table \ref{tab:stats1}. 
The total number of test instances for both Spanish and Chinese languages exceed 150,000. 
\begin{table}[h]
    \centering
    \begin{tabular}{c|c|c}
          & \textbf{Train} & \textbf{Dev} \\
          \hline
         \# sentences & 15300 & 800 \\
    \end{tabular}
    \caption{Total sentences in Chinese and Spanish monolingual track}
    \label{tab:stats1}
\end{table}

\section{System Overview}
\label{method}
At first, we pre-train the BERT language model on unlabelled corpora for the target low resource language. For Chinese, we use the strategy outlined by \cite{cui-etal-2020-revisiting}. BERT uses the WordPiece tokenizer \cite{wordpiece} to split tokens into smaller fragments. It is easier for the masked language model to predict these masked fragments. However, for the Chinese textual texture, the Chinese characters are not formed by alphabet-like symbols, so the WordPiece tokenizer is unable to split the words into small fragments. Hence, we use the Chinese Word Segmentation (CWS) tool to split the text into separate words and then use Whole Word Masking (WWM) strategy for the masked language model objective. 
In comparison to masking small fragments, this Whole Word Masking strategy makes it harder for the model to predict whole masked words, leading to more robustness.

For the Spanish variant, we adopt the strategy outlined by \citet{spanish}. Similar to \cite{cui-etal-2020-revisiting}, they use the strategy of whole word masking for pre-training BERT language model on unlabelled Spanish corpus.

We adopt the strategy of finetuning these pre-trained BERT models on the downstream NER task for each language.

\subsection{BERT+CRF} 
\begin{figure}[t!]
  \centering
  \includegraphics[width=0.95\linewidth]{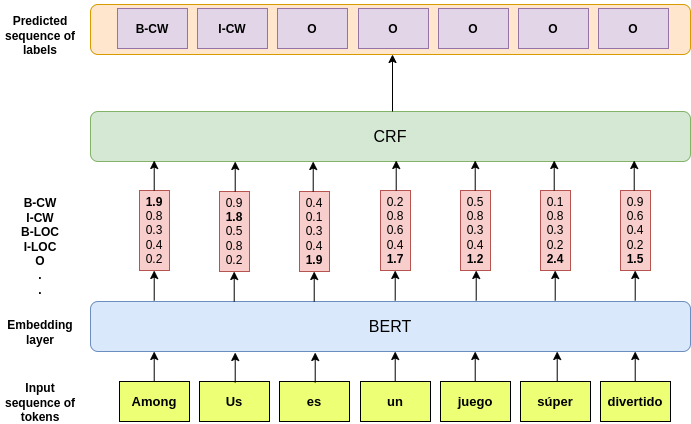}
  \caption{BERT+CRF architecture}
  \label{fig:BERT-CRF architecture}
\end{figure}

Conditional random fields (CRFs) are statistical modeling methods used for pattern recognition. They are better suited for tasks such as Part-of-Speech (POS) tagging and NER compared to classifiers based on softmax normalization. Classifiers based on softmax normalization assume the likelihood of the labels to be conditionally independent, and this causes label bias. CRF alleviates this issue of label bias by capturing inter-token dependencies in a graphical model and learning transition scores in addition to the hidden states. 
In our model, we use linear-chain CRF. In linear CRFs, prediction for each token is dependent only on its immediate neighbors. As shown in equation \ref{equation:eq2}, CRF tries to maximize the ratio of the probability of an optimal sequence of labels to the probability sum of all the possible sequences of labels. Since CRF focuses on the sequence of labels, it can avoid errors like B-PER followed by an I-PROD. Therefore, based on emission scores provided by BERT layer, we calculate the log-likelihood of a sequence of labels. Now we explain the steps involved BERT+CRF architecture that is shown in Figure \ref{fig:BERT-CRF architecture}.

Firstly, we obtain token-level dense representations using a fine-tuned BERT-based embedding layer.
For an input sequence of tokens $w = (w_{1},w_{2},w_{3},...,w_{n})$, we obtain the $i$th token representation $x_{i}$ of dimension $d$, where $d$ is the dimension of BERT embeddings.
The token embedding $x_{i}$ is passed to a dense linear layer to transform the representation from $d$ to $k$ dimensional space (emission score), where $k$ is the number of labels. We calculate emission scores for all the tokens of the given sequence.
We then pass these emission scores to the CRF layer to obtain the probability for a sequence of labels. The emission scores, obtained from the previous layer as $P \in R^{n\times{k}}$, are passed to the CRF layer whose parameters are $A \in R^{k+2\times{k+2}}$. Element $A_{ij}$ denotes the transition score from the $i$th to the $j$th label. 2 additional states are added to the start and end of the sequence.  For a series of tokens $w = (w_{1},w_{2},w_{3},...,w_{n})$ we obtain a series of predictions $y = (y_{1},y_{2},y_{3},...,y_{n})$. 


As described in \cite{lample-etal-2016-neural}, the score of the entire sequence is defined as :


\begin{equation}
 s(w,y) = \sum_{i=0}^{n}A_{y_{i},y_{i+1}} + \sum_{i=1}^{n}P_{i,y_{i}}
 \label{equation:eq1}
 \end{equation}

The model is trained to maximise the log probability of the correct label sequence:
\begin{equation}
    \log(p(y|w))= s(w,y)-\log(\sum_{\tilde{y}\in \textbf{$Y_w$}}e^{s(w,\tilde{y})})
    \label{equation:eq2}
\end{equation}


where \textbf{$Y_w$} are all possible label sequences.

\subsection{BERT+BiLSTM+CRF} We obtain token-level contextual dense representations (BERT embeddings) using fine-tuned BERT layer. These embeddings are then passed to a BiLSTM layer which further extracts bidirectional information from the given sequence of vectors. The information is encoded in the hidden-state representations of the BiLSTM. We pass these hidden states to the CRF layer to obtain the likelihood of a sequence of labels.
We use the pre-trained language model to map the tokens in each input sentence to a dense embedding representation. The BERT-based dense embeddings are passed to the BiLSTM-CRF layer, which is used to obtain the predicted label for each token in the entire sequence. More formally, for a sequence of tokens $w = (w_{1},w_{2},w_{3},...,w_{n})$, we obtain the $i$th token representation $x_{i}$ of dimension $d$, which is the dimension of the dense vector representations of the BERT-based embeddings obtained from the pre-trained language model. The sequence of token embeddings is taken as an input to the BiLSTM in each time step, and the forward hidden states $\overrightarrow{h_{f}} = (\overrightarrow{h_{1}},\overrightarrow{h_{2}},\overrightarrow{h_{3}},...,\overrightarrow{h_{n}})$ and the backward hidden states $\overleftarrow{h_{b}} = (\overleftarrow{h_{1}},\overleftarrow{h_{2}},\overleftarrow{h_{3}},...,\overleftarrow{h_{n}})$ are concatenated to form the combined hidden state representation $h = [\overrightarrow{h_{f}},\overleftarrow{h_{b}}]$. The combined hidden state representation $h \in R^{n\times{m}}$, where $m$ is the total dimension of BiLSTM, is transformed to a $k$ dimensional space using a linear layer, where $k$ is the total number of labels. Finally, the CRF layer outputs predicted sequence of labels.

\subsection{BERT+Linear} This is the simplest architecture based on fine-tuned BERT layer. The input token sequence is mapped to a vector space of $d$ dimension using a pre-trained BERT layer. These embeddings are then passed to a classifier that consists of two Fully Connected (FC) layers followed by a softmax normalization function. The classifier maps the $d$ dimensional BERT embeddings to $k$ dimensions, where $k$ is the number of labels. These $k$ dimensional vectors generated by the fully connected layers are softmaxed to provide a probability distribution across all labels.


\begin{table*}[]
    \centering
    \begin{tabular}{c|c|c|c|c|c|c|c|c|c|c|}
          & \multicolumn{3}{|c|}{\textbf{BERT+CRF}} & \multicolumn{3}{|c|}{\textbf{BERT+BiLSTM+CRF}}& \multicolumn{3}{|c|}{\textbf{BERT+Linear}}\\
         \hline
    Class Label & Prec & Rec & F1 & Prec & Rec & F1 & Prec & Rec & F1\\
    \hline
    LOC &  0.8368 & 0.8796 & \textbf{0.8577} & 0.8219 & 0.8759 & 0.8481 & 0.8194 & 0.8613 & 0.8399  \\
    PER &  0.9065 & 0.9028 & 0.9047 & 0.9177 & 0.9028 & \textbf{0.9102} & 0.8933 & 0.9150 & 0.9040  \\
    PROD & 0.6970 & 0.7468 & 0.7210 & 0.7278 & 0.7468  & \textbf{0.7372}  & 0.6864 & 0.7532 & 0.7183 \\
    GRP & 0.7952 & 0.7857 & 0.7904 & 0.7751 & 0.7798 & 0.7774 & 0.8061 & 0.7917 & \textbf{0.7988} \\
    CW & 0.7965 & 0.7135 & 0.7527 & 0.7654 & 0.7135 & 0.7385 & 0.8107 & 0.7135 & \textbf{0.7590} \\
    CORP & 0.8657 & 0.8227 & \textbf{0.8436}  & 0.8397 & 0.7801 & 0.8088 & 0.8529 & 0.8227 & 0.8375\\
    \hline
    Average & 0.8163 & 0.8085 & \textbf{0.8117} & 0.8079 & 0.7998 & 0.8034 & 0.8115 & 0.8096 & 0.8096 \\
    \end{tabular}
    \caption{Results of our models on validation dataset for Spanish language}
    \label{tab:results1}
\end{table*}

\begin{table*}[]
    \centering
    \begin{tabular}{c|c|c|c|c|c|c|c|c|c|c|}
          & \multicolumn{3}{|c|}{\textbf{BERT+CRF}} & \multicolumn{3}{|c|}{\textbf{BERT+BiLSTM+CRF}}& \multicolumn{3}{|c|}{\textbf{BERT+Linear}}\\
         \hline
    Class Label & Prec & Rec & F1 & Prec & Rec & F1 & Prec & Rec & F1\\
    \hline
    LOC & 0.9465 & 0.9365 & \textbf{0.9415} &  0.9239 & 0.9312 & 0.9275 &  0.9186 & 0.9259 & 0.9223  \\
    PER &  0.8497 & 0.9225 & 0.9084 & 0.8971 & 0.9457 & \textbf{0.9208} &  0.8955 & 0.9302 & 0.9125  \\
    PROD &  0.8867 & 0.8285 & 0.8566 & 0.8662 & 0.8504 & \textbf{0.8582} &  0.8593 & 0.8248 & 0.8417  \\
    GRP &  0.7500 & 0.6923 & \textbf{0.7200} & 0.7727 & 0.6538 & 0.7083 &  0.6923 & 0.6923 & 0.6923  \\
    CW &  0.8265 & 0.8617 & \textbf{0.8437} & 0.8556 & 0.8191 & 0.8370 &  0.8370 & 0.8191 & 0.8280  \\
    CORP & 0.8615 & 0.8750 & 0.8682 & 0.8808 & 0.8854 & \textbf{0.8831} & 0.8883 &  0.8698 & 0.8789 \\
    \hline
    Average &  0.8610 & 0.8527 & \textbf{0.8564} & 0.8660 & 0.8476 & 0.8558 &  0.8485 & 0.8437 & 0.846 \\
    \end{tabular}
    \caption{Results of our models on validation dataset for Chinese language}
    \label{tab:results2}
\end{table*}

\begin{table*}[h]
    \begin{minipage}{0.48\textwidth}
        \centering
        \begin{tabular}{p{3.5cm}|p{0.75cm}|p{0.75cm}|p{0.75cm}}
        & \textbf{Prec} & \textbf{Rec} & \textbf{F1}  \\
        \hline
        Baseline System &  0.764 & 0.763 & 0.767\\
        BERT + Linear & 0.811 & \textbf{0.809} & 0.809\\
        BERT+BiLSTM+CRF & 0.807 & 0.799  & 0.803 \\
        BERT + CRF & \textbf{0.816} & 0.808 & \textbf{0.811} \\
    \end{tabular}
        \caption{Comparison of models' performances with baseline on validation dataset for Spanish language} 
    \label{tab:results3}
    \end{minipage}
    \hfill
    \begin{minipage}{0.48\textwidth}
        \centering
        \begin{tabular}{p{3.5 cm}|p{0.75cm}|p{0.75cm}|p{0.75cm}}
         & \textbf{Prec} & \textbf{Rec} & \textbf{F1}  \\
        \hline
        Baseline System & 0.758 & 0.762 & 0.755\\
        BERT + Linear & 0.848 & 0.843  &0.846 \\
        BERT+BiLSTM+CRF & \textbf{0.866} & 0.847 & 0.855 \\
        BERT+CRF &  0.861 & \textbf{0.852} & \textbf{0.856}\\
    \end{tabular}
    \caption{Comparison of models' performances with baseline on validation dataset for Chinese language} 
    \label{tab:results4}
    \end{minipage}
\end{table*}

\begin{table*}[h]
    \begin{minipage}{0.48\textwidth}
        \centering
    \begin{tabular}{c|c|c|c}
          & \multicolumn{3}{|c}{\textbf{BERT+CRF}}\\
         \hline
        Class Label & Prec & Rec & F1 \\
        \hline
        LOC & 0.5768 & 0.6571 & 0.6144\\
        PER & 0.7641 & 0.7739 & 0.7690\\
        PROD & 0.6292 & 0.5141 & 0.5659\\
        GRP & 0.5727 & 0.5560 & 0.5642\\
        CW & 0.5331 & 0.5257 & 0.5294\\
        CORP & 0.6605 & 0.6005 & 0.6291\\ 
        \hline
        Average & 0.6227 & 0.6046 & 0.6120\\
        \end{tabular}
        \caption{Performance of Spanish model on test dataset}
        \label{tab:results7}
    \end{minipage}
    \hfill
    \begin{minipage}{0.48\textwidth}
        \centering
        \begin{tabular}{c|c|c|c}
                  & \multicolumn{3}{|c}{\textbf{BERT+CRF}}\\
                 \hline
            Class Label & Prec & Rec & F1 \\
            \hline
            LOC & 0.6930 & 0.7955 & 0.7407\\
            PER & 0.7952 & 0.6377 & 0.7078\\
            PROD & 0.6853 & 0.7232 & 0.7038\\
            GRP & 0.7254 & 0.4608 & 0.5636\\
            CW & 0.5520 & 0.6798 & 0.6093\\
            CORP & 0.6526 & 0.7361 & 0.6918\\ 
            \hline
            Average & 0.6839 & 0.6722 & 0.6695\\
            \end{tabular}
            \caption{Performance of Chinese model on test dataset}
            \label{tab:results8}
    \end{minipage}
\end{table*}

\section{Implementation Details}
\label{implementationdetails}
We implement all our transformer based models using Pytorch and Huggingface library. The Chinese language model with the Whole Word Masking (WWM) objective is trained on the Chinese Wikipedia unlabelled text corpus. We use the same training corpus of 3 billion unannotated Spanish tokens as \citet{spanish} to pre-train the BERT language model on Spanish data.
We implement 3 models: BERT+CRF, BERT+BiLSTM+CRF, and BERT+Linear, for our low resource NER task setting. We run our experiments between 1-100 epochs. We find that the best results are obtained at 10 epochs of training for each model after which the model starts to overfit. We use a dropout from 0.2 to 0.5 for all models. We employ Adam optimizer with default parameters for all experiments. We also experiment with a cyclic learning rate between ${1e^{-4}}$ to ${1e^{-6}}$ to avoid getting stuck in local minima. The size of each of the FC layers in the BERT+Linear model is 512. We validate the results of all models using our validation dataset. The hidden layer size of BiLSTM used in the BERT+BiLSTM+CRF model is 256.

\section{Results}
\label{results}
We compare the performances of all models in the low-resource setting for both Chinese and Spanish languages. From tables \ref{tab:results1} and \ref{tab:results2} we observe that the BERT+CRF model performs the best across both languages on validation set. We choose the best performing model to evaluate our results on the blind test set. 
The baseline model chosen by the organisers of this task is XLM-RoBERTa \cite{xlm-r}(base model). It is pre-trained on 2.5TB of filtered CommonCrawl data containing 100 languages \cite{xlm-r}.
Our approach beats the baseline by a significant margin and outperforms multiple models in the competition. We present the precision, recall and F1 scores for all 3 models in the Tables \ref{tab:results1} and \ref{tab:results2} for Spanish and Chinese language respectively. We also compare the results between the baseline and our models for the validation dataset in the Tables \ref{tab:results3} and \ref{tab:results4}.

For the Spanish language, we observe that the BERT+CRF (0.8117 F1 score) beats BERT+Linear (0.8096 F1 score) by a slender margin. This can be attributed to the addition of the CRF layer, which exploits inter-token dependencies. The BERT+BiLSTM+CRF model is much heavier with a larger number of parameters and overfits the training dataset due to the smaller number of training instances. 

For the Chinese language, we observe that both the BERT+CRF (0.8564 F1 score) and BERT+BiLSTM+CRF (0.8558 F1 score) beat BERT+Linear (0.846 F1 score).

Our models outperform the baseline for both the languages. Our best performing model BERT+CRF beats the baseline F1 score by around 5$\%$ for Spanish and by around 10$\%$ for Chinese.

The details of the performance of the BERT+CRF model in the evaluation phase are shown in Table \ref{tab:results7} for the Spanish language and the Table \ref{tab:results8} for the Chinese language. We observe a drop in performance on the test dataset compared to the performance on the validation dataset. The model scores 0.6120 F1 for Spanish and 0.6695 F1 for the Chinese language.

\section{Error Analysis}
\label{erroranalysis}
We perform error analysis on all 3 different models. We qualitatively analyze the predictions on the validation dataset for both languages. As the final evaluation test set in blind, we are unable to perform analysis on the same. We find that the labels GRP (Group), PROD (Product), and CW (Creative Work) are the most inaccurately predicted labels for the Spanish models. This conforms to our hypothesis that the long-tailed nature of these entities (which means the frequency of occurrence of such entity types in the general literature of the target language is rare). Hence, the model has the most difficulty in recognizing these entities from the contextual sentences. The other label types are more common and were present in the CoNLL dataset as well. We also notice that the BERT+Linear does marginally better than BERT+CRF on predicting such labels (for the Spanish language), despite it not being the best performing model overall. This can be attributed to it being a lighter model, imparting it the capability of generalizing better while training on a relatively lower amount of training instances.
BERT+CRF benefits from having CRF along with the lower number of parameters compared to the BERT+BiLSTM+CRF model. This results in it having a better performance compared to both the other models.
The drop in the performance of the model on the blind test dataset can be attributed to the model not generalizing well to handle instances of questions and short search queries in the additional test set.

\section{Conclusion and Future Work}
\label{conclusion}
We have introduced strong improvements over the baseline for the shared task of complex NER for low resource languages. We leverage the Whole Word Masking objective to obtain a better performance in this low-resource setting. We perform extensive experiments and find that simple BERT-CRF based models perform strongly against other heavier models even in such low resource semantically ambiguous setting as evident by the final evaluation rankings. We find this approach to give a higher performance as it is able to utilize the contextual information from a sequence of tokens and learn inter-token dependencies to accurately predict the named entity labels.  We also conduct qualitative error analysis and describe our findings. For future work, we aim to leverage these findings to circumvent the label scarcity problem in other low-resource languages and code mixed data.

\section{Acknowledgment}
We thank the SemEval team, the task organizers, and other participants of this shared task.

\bibliography{anthology,custom}
\bibliographystyle{acl_natbib}

\end{document}